\documentclass[10pt,twocolumn,letterpaper]{article}

\usepackage{cvpr}
\usepackage{times}
\usepackage{epsfig}
\usepackage{graphicx}
\usepackage{amsmath}
\usepackage{amssymb}

\usepackage[breaklinks=true,bookmarks=false]{hyperref}

\cvprfinalcopy 

\def\cvprPaperID{****} 

\begin{document}

\title{A Mask-RCNN Baseline for Probabilistic Object Detection}

\author{Phil Ammirato\\
UNC-Chapel Hill\\
\tt\small ammirato@cs.unc.edu\\
\and
Alexander C. Berg\\
UNC-Chapel Hill\\
\tt\small aberg@cs.unc.edu
}

\maketitle

\begin{abstract}

The Probabilistic Object Detection Challenge evaluates object detection methods using a new evaluation measure, Probability-based Detection Quality (PDQ), on a new synthetic image dataset. We present our submission to the challenge, a fine-tuned version of Mask-RCNN with some additional post-processing. Our method, submitted under username pammirato, is currently second on the leaderboard with a score of 21.432, while also achieving the highest spatial quality and average overall quality of detections. We hope this method can provide some insight into how detectors designed for mean average precision (mAP) evaluation behave under PDQ, as well as a strong baseline for future work. 

\end{abstract}

\section{Introduction}

\footnote{Supported by NSF NRI grant 1526367}

Object detection is a both very popular task in the computer vision research community and a useful tool for many real world applications. Many such applications are in robotics. A robot aided with information of the objects around it will be better equipped to have successful interactions with its environment. 

Traditionally, object detection methods are designed and evaluated for the Mean Average Precision (mAP) metric. This metric is very useful for evaluating detection methods for a variety of tasks, but may not be ideal for robotics contexts. mAP ranks the bounding boxes output by a detection system based on each box's score. The raw values of the scores are hard to interpret, other than a higher score is means the system is more confident. The ranking system generally encourages the detectors to output many detections for each image, as usually this will only increase its mAP score, as shown in \cite{pdq}.  This large output may not be ideal for a robotics context, as processing resources may be limited, and there is no reliable/consistent measure for confidence. 

Recently, a new evaluation technique for object detectors was proposed, the Probability-based Detection Quality (PDQ) \cite{pdq}. This measure treats detection scores as real probabilities, not just as means for a ranking mechanism. It also allows for some uncertainty to be modeled in the objects location, represented as a \emph{probabilistic bounding box} (pbox). A covariance matrix is given for each corner of a pbox, describing the uncertainty of the box's shape. This gives a much more felixble representation that a traditional bounding box, which is very rigid in what is an object and what is not.

The goal of this work is to establish a strong baseline for probabilistic object detection using an existing well known object detector, Mask-RCNN \cite{he2017maskrcnn}. We first fine-tune a Mask-RCNN model trained for MSCOCO \cite{lin2014microsoftCOCO} mAP object detection for the detection on the PODC data. We then develop some post-processing routine on the detectors output to be better suited to PDQ evaluation as opposed to mAP. We tested our work as a part of the Probablistic Object Detection Challenge (PODC) at CVPR 2019 and achevied 2nd place.

\section{Mask-RCNN}
Mask-RCNN \cite{he2017maskrcnn} is a very popular deep-learning method for object detection and instance segmentation that achieved state-of-the art results on the MSCOCO\cite{lin2014microsoftCOCO} dataset when published. While a few detectors have since passed Mask-RCNN in mAP performance, they have done so by only a few points and are usually based on the Mask-RCNN architecture. 

Mask-RCNN is a two-stage recognition pipeline. In its first stage, features are extracted from the image using a backbone convolutional neural network (CNN), and class agnostic region proposals are predicted. These proposals are then refined and classified in the second stage, to become either labeled bounding boxes for object detection or segmentation masks for the instance segmentation task. 

We use an implementation of Mask-RCNN based in Pytorch \cite{pytorch2017}, \cite{maskrcnn_benchmark_code}. The authors of \cite{maskrcnn_benchmark_code} provide models trained for object detection and instance segmentation on MSCOCO, using backbone networks that are pre-trained on the ImageNet \cite{russakovsky2015imagenet} classification dataset. Their is also code for training models from scratch or fine-tuning.  See their public github repository for more details.  We base our system for the PODC challenge on the models provided by this code base.

\section{Training for PODC data}

We started building our system using a model from \cite{maskrcnn_benchmark_code} that used a ResNet-101 \cite{resnet} backbone with a Feature Pyramid Network \cite{feature_pyramid}, and was trained for both object detection and instance segmentation on the 80 object classes of MSCOCO. This model achieves 42.2 mAP on the 80 MSCOCO classes. Our first step was to remove all mask prediction heads and the object detection classifier heads for the 50 classes in MSCOCO that are not included in PODC. This left us with an object detector for 30 classes designed and trained to maximize mAP on MSCOCO, we will refer to this as Mask-RCNN-30. 

The lack of training data for this task presents a unique domain transfer challenge. Usually systems aim to transfer from synthetic images to real world images, but in this task the test data is synthetic while there is some real training data available. 

The PODC validation and testing data is generated from a high-fidelity simulation, and so the images are likely from a different domain than the real-world images collected in MSCOCO. To see how well our initial model can generalize to this new domain, we test test Mask-RCNN-30 on one scene of the PODC validation data, which will refer to as PODC-val0. As shown in Table \ref{tab:training_mAP}, the model trained only on MSCOCO does not perform well on PODC-val0 when compared to its performance on MSCOCO. 

Since there is no training data associated with the PODC, we looked for another source of synthetic data to help Mask-RCNN-30 generalize better. We first looked at the AI2-Thor \cite{ai2thor} environment, another synthetic renderer built for simulating robotic motion around indoor home-style scenes. Unfortunately, the object class set of AI2-Thor does not fully cover the set of objects in PODC, only about 9 of PODC classes are labeled. In fact there are object classes, such as oven, that are present in the images of AI2-Thor but are not labeled. These false negatives in the ground truth data would likely cause confusion during training. 

The SunCG \cite{suncg} dataset, however, has a much higher overlap of labeled classes with PODC with 25 of the 30 classes covered. House3D \cite{house3d} uses the 3D scene models and annotations to provide environments similar to AI2-Thor. We use tools from the publicly available House3d code to generate a training set of 173,250 images with bounding box annotations, which we refer to as House3D-train.

While the SunCG data is synthetic, it is much less realistic than the PODC validation and testing data. We fine-tune Mask-RCNN-30 on both the original MSCOCO images it was trained on and the new House3D-train set, in a weak attempt at domain randomization with respect to image source. The hope is that the model will learn to be robust to different image sources, real or synthetic. We can see in Table \ref{tab:training_mAP} that this does help performance on PODC-val0 with respect to mAP.

We add one more data augmentation during training, namely adjusting the image brightness and contrast via Pytorch transforms. There are two intended effects of this augmentation: to further improve the generalization of the model to different image source domains, and to help with performance on the `night' scenes present in PODC. We can see another small improvement in mAP in Table \ref{tab:training_mAP}.

In Table \ref{tab:training_mAP}, we can see as we improve mAP, PDQ scores actually decrease. This is mostly due to an increase in false positives, and we will give some heursitcs for remedying this in the next section.

\begin{table}
    \centering
    \begin{tabular}{c|c|c}
       Model  &  mAP & PDQ \\
       \hline
        Mask-RCNN-30& 12.09 & 7.2 \\
        MSCOCO + SunCG & 13.53 & 6.16 \\
        MSCOCO + SunCG + jitter & 14.08 & 5.73 \\
    \end{tabular}
    \caption{mAP and PDQ on PODC-val0 for different training data. mAP improves as we add more variation to the training data. PDQ gets worse mainly due to an increase of false positives. }
    \label{tab:training_mAP}
\end{table}

\begin{figure*}
    \centering
    \includegraphics[width=0.8\textwidth]{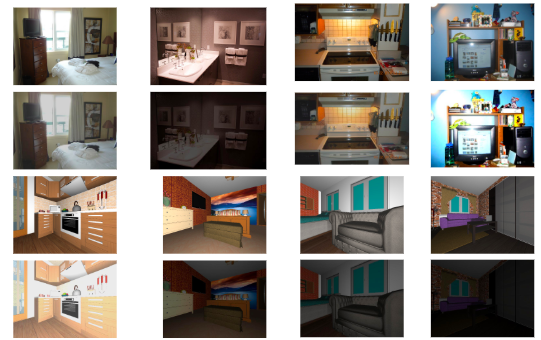}
    \caption{In each respective row, training data from: MSCOCO\cite{lin2014microsoftCOCO}, MSCOCO + jitter, SunCG-House3D, SunCG-House3D + jitter. }
    \label{fig:train_data}
\end{figure*}
\begin{figure*}
    \centering
    \includegraphics[width=0.8\textwidth]{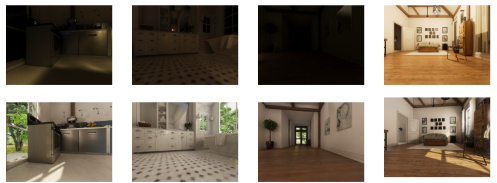}
    \caption{Validation images from PODC. }
    \label{fig:podc_val}
\end{figure*}

\section{Post Processing for PODC evaluation}

Even though we have managed to improve our detectors mAP score on PODC-val0, the PDQ score is still very low and has been getting worse. We identify various causes of our low PDQ score, and propose simple post-processing techniques to improve them. These improvements are not meant to solve the problem of adapting detectors for probabilistic object detection. They hopefully provided some insights into the problem, and will serve as a useful baseline for future techniques. 

\subsection{PDQ Calculation}
We first give a short outline of how PDQ \cite{pdq} is calculated, so the motivations for our solutions will be clear. 

There are two important two sub-measures of PDQ we must first compute, spatial quality ($Q_{S}$) and label quality ($Q_{L}$). 

Let $T_{j}^{i}$ be the probability distribution across all possible object classes assigned to the $j^{th}$ in the $i^{th}$ image. Then the label quality of the detection, $D_{j}^{i}$ with respect to the $k^{th}$ ground truth object, $G_{k}^{i}$ is: 

\begin{equation}
Q_{L} (G_{k}^{i}, D_{j}^{i})=  T_{j}^{j}(c_{k}^{i})
\end{equation}

where $c_{k}^{i})$ is the object class of $G_{k}^{i}$. This is a number on $[0,1]$, that is equal to the detector's score for the ground truth object, regardless of any other detections or objects.

The spatial quality, $Q_{S}$, is calculated based on the ground truth's objects segmentation mask and the probabilistic bounding box outputted by the detector. Essentially, assigning higher probabilities to pixels that belong to the object as foreground improve the score, and including any background pixels outside of the ground truth bounding box hurt the score.

The pairwise PDQ for one detection and ground truth is then calculated using the geometric mean of these two measures. 
\begin{equation}
pPDQ(G_{k}^{i}, D_{j}^{i}) = \sqrt{Q_{S} * Q_{L}}
\end{equation}

A matching routine then finds the best matching of detections and ground truth objects. The average score after the matching routine for all true positive detections is $aPDQ$.  the final PDQ score is calculated. The score is scaled by three quantities: the number of false positives, $N_{fp}$, the number of false negatives, $N_{fn}$ and the number of true postives, $N_{tp}$.

\begin{equation}
PDQ =  \frac{1}{N_{fp} + N_{fn} +N_{tp}} * N_{tp}*aPDQ 
\end{equation}

\subsection{False Negatives}

Perhaps the largest source of error when evaluating raw Mask-RCNN output with PDQ is the large number of false negatives. 
The simple solution is to remove all detections with score less than .5. We experimented with changing this score threshold, but .5 consistently gave the best performance as show in Table \ref{tab:training_thresh}.

\begin{table}
    \centering
    \begin{tabular}{c|c|c|c|c}
        Score Threshold  & PDQ & TP & FP & FN \\
       \hline
         0.0 & 5.73 & 7262 & 41771 & 7978 \\
          0.3 & 14.78 & 5331 & 6412 & 9909 \\
        0.4 & 15.58 & 4994 & 4155 & 10246 \\
        \textbf{0.5} & \textbf{15.79} & 4694 & 2875 & 10546 \\
        0.6 & 15.63 & 4296 & 1988 & 10944 \\
        0.7 & 15.21 & 3915 & 1210 & 11325 \\
    \end{tabular}
    \caption{How PDQ score changes when changing the score threshold for discarding detections. This includes the heuristic of setting all detection scores to 1.0 and a set covariance of 7.5. TP=True Positives, FP=False Positives, FN=False Negatives.   }
    \label{tab:training_thresh}
\end{table}

\subsection{Label Quality}
The label quality is only counted for true positive detections. The only penalty for assigning high scores to false positive detections comes from the spatial quality measure. We found that given the large weight of label quality in the final PDQ score, and the other influences on spatial quality besides detection score, reassigning a value of 1.0 to all detection scores resulted in the highest PDQ. This seems to not be in the spirit of PDQ as now the scores have even less meaning than the original detector output, while PDQ seems to encourage a meaningful uncertainty measure. We hope that with this baseline established, future work will have to find some better representation for detection scores to improve.

\subsection{Confusing Objects}\label{sec:confuse}
While removing all low scoring detections was a simple and effective solution for reducing false positives, it also removed a lot of false positives. We added another simple heuristic to add back in certain low scoring detections that had boxes with high intersection over union (IOU) with other low scoring detections. Essentially here we are trying to add in detections that are likely objects of interest, as the detector has at least two outputs for the object, but have ambiguous class to the detector. 

As can be seen in the equation 3, while $N_{fn}$ >0, adding a true positive detection has improve the score more than it will be reduced from adding a false positive. So adding keeping two detections, even though one is almost guaranteed to be a false positive, is worth it as long as one is a true positive. This method increases the performance only a small amount, and is likely not worth the expensive computation.

\subsection{Probabilistic Bounding Box}
We found that adding some fixed value in the covariance matrix to our bounding boxes was slightly better than nothing, but scaling the covariance according to the box size was even better. After trying various values, adding a covariance between 20-30\% of the bounding box size was best. For example, scaling at 30\%, a bounding box with width of 100 and height of 50 would be given a covariance matrix of:
$$
\begin{bmatrix}
    30.0 & 0.0  \\
    0.0 & 15.0  \\
\end{bmatrix}
$$

for both the top left and bottom right corners. See Table \ref{tab:training_cov} for an ablation study.

\begin{table}
    \centering
    \begin{tabular}{c|c|c|c|c}
        Covariance  & PDQ & Avg Spatial \\
       \hline
        0.0 & 6.51 & 17.50 \\
        5.0 & 15.45 & 44.26  \\
        7.5 & 15.79 &  45.04 \\
        10.0 & 15.97 &  45.26 \\
        15.0 & 16.12 &  45.17 \\
        20.0 & 16.13 &  44.75 \\
        25.0 & 16.08 &  44.19 \\
        10\% & 16.06 &  45.89 \\
        \textbf{20\%} & \textbf{16.36} &  \textbf{46.05} \\
    \end{tabular}
    \caption{How PDQ score changes when changing how to set the covariance of every bounding box. Compares both setting the covariance to a set number (i.e. 5.0) and as a function of the box dimensions (i.e. 20\% box). Includes heuristics of removing detections with scores below 0.5 and then setting all detection scores to 1.0. \emph{Avg Spatial} is the average spatial quality as defined by the evaluation code provided by the PODC. PDQ goes up slightly when covariance goes from 10.0 to 15.0 even though spatial quality decreases, because the number of True Positives also changes slightly.   }
    \label{tab:training_cov}
\end{table}

\subsection{Reducing Box Size}
The spatial quality measure can be reduced greatly by assigning high probabilities to pixels that are truly in the background. Since all of our detections have a score of 1.0, this can really hurt our overall PDQ. We reduce the bounding box size of all detections by 10\% in width and height respectively. This gives us a smaller box with high confidence centered on each object, and our large covariance values allow us to capture the rest of the foreground pixels without such a high probability assigned to the background. 

\begin{table}
    \centering
    \begin{tabular}{c|c|c|c|c}
        Box Reduction Factor  & PDQ & Avg Spatial \\
       \hline
        0.0 & 16.36 & 46.05 \\
        \textbf{0.1} & 18.11 & 55.00  \\
        0.2 & 14.72 &  37.76 \\
    \end{tabular}
    \caption{How PDQ score changes when reducing the size of every bounding box. Includes heuristics of removing detections with scores below 0.5 and then setting all detection scores to 1.0, as well as setting the covariance as 20\% of the reduced box size.  \emph{Avg Spatial} is the average spatial quality as defined by the evaluation code provided by the PODC.   }
    \label{tab:training_boxsize}
\end{table}

\section{Final Model Details}

For our final model, we fine-tune Mask-RCNN-30 on the MSCOCO + SunCG + jitter training set for 10,000 iterations at a batch size of 8 and learning rate or .0005. We then add the PODC validation data to the training set and fine-tune for another 5,000 iterations. We post process the detections by thresholding at a score of 0.5, then setting all scores to 1.0. We then add back in boxes as in section \ref{sec:confuse}. Finally we reduce the bounding boxes by 10\% and add in a covariance based on 30\% of the box width and height.

\section{Conclusion and Future Work}
 With a final score of 21.432, we see lots of room for improvement in future work, though the lack of training data is a limiting factor. Most of the improvements in this work are based on simple heuristics taking advantage of the structure of the PDQ calculation.  We hope future work can use this to improve, and provide more meaningful uncertainty predictions.

{\small
\bibliographystyle{ieee_fullname}
\bibliography{final}
}

\end{document}